# Dynamic Mask-Based Backdoor Attack Against Vision AI Models: A Case Study on Mushroom Detection

Zeineb Dridi[1], Jihen Bennaceur[2,3], and Amine Ben Hassouna[2,4]

[1] Oakland University, Department of Computer Science and Engineering, Rochester MI 48309, USA
[2] Mediterranean Institute of Technology, South Mediterranean University, Tunis, 1053, Tunisia
[3] National School of Information Science (ENSI), University of Manouba, Manouba, 2010, Tunisia
[4] Dracodes, Tunis, 1006, Tunisia

**Abstract.** Deep learning has revolutionized numerous tasks within the computer vision field, including image classification, image segmentation, and object detection. However, the increasing deployment of deep learning models has exposed them to various adversarial attacks, including backdoor attacks. This paper presents a novel dynamic mask-based backdoor attack method, specifically designed for object detection models. We exploit a dataset poisoning technique to embed a malicious trigger, rendering any models trained on this compromised dataset vulnerable to our backdoor attack. We particularly focus on a mushroom detection dataset to demonstrate the practical risks posed by such attacks on critical real-life domains. Our work also emphasizes the importance of creating a detailed backdoor attack scenario to illustrate the significant risks associated with the outsourcing practice. Our approach leverages SAM, a recent and powerful image segmentation AI model, to create masks for dynamic trigger placement, introducing a new and stealthy attack method. Through extensive experimentation, we show that our sophisticated attack scenario maintains high accuracy on clean data with the YOLOv7 object detection model while achieving high attack success rates on poisoned samples. Our approach surpasses traditional methods for backdoor injection, which are based on static and consistent patterns. Our findings underscore the urgent need for robust countermeasures to protect deep learning models from these evolving adversarial threats.

## 1 Introduction

Deep Neural Networks (DNNs) have significantly advanced the field of Artificial Intelligence (AI), achieving remarkable performances in computer vision tasks such as image classification [1], object detection [2], and image segmentation [3]. Their ability to extract intricate patterns and representations from complex datasets has fundamentally transformed how machines process and comprehend visual information. These capabilities allowed DNNs to achieve exceptional endeavors across various tasks and made them invaluable in modern mission-critical applications such as face recognition [4], autonomous vehicles [5], surveillance [6], and X-ray analysis [7]. However, with these advancements also emerged significant security concerns that threaten the integrity and reliability of AI systems. One example of these threats is adversarial attacks where an adversary introduces manipulations to input to fool the AI model into causing misclassification or degrade its performance [8] [9] [10] [11]. Among these threats, are backdoor attacks which are significantly raising concerns due to their high stealthiness and effectiveness. In this type of attack, the adversary embeds hidden behavior within the DNN models during the training phase. This malicious behavior occurs only when triggered by specific inputs. Therefore, the backdoor model continues to behave normally on clean samples while causing incorrect outputs when encountering triggering inputs carrying a specific pattern of trigger.

The earliest work on backdoor attacks was published in 2017 [12], where the authors demonstrate the effectiveness of these attacks and their consequences on mission-critical applications such as autonomous driving. Since then, an increasing number of works have been devoted to the study of these attacks with the intention of enlarging the attack methods and techniques proposed by earlier works. Our paper presents a novel contribution to this field by introducing a new stealthy and effective technique for backdoor attacks.

To be more specific our contributions to the AI security research community are the following:

- We provide a summary and survey on existing attack methods and their limitations, especially in the design of the backdoor trigger. Our work will explore the benefits and the limitations of these methods in order to propose a new attack technique.
- We propose a realistic attack scenario where a backdoor attack can be implemented to demonstrate the process behind the injection of these attacks within DNNs. Our project considers a specific real-world case study of a backdoor attack on a mushroom detection model that could lead to critical health issues within a whole community. In this scenario, we propose new concepts including personas and timelines to make a more detailed and thorough description of the individuals and steps involved in such an attack.
- We introduce our novel attack method that leverages the latest image segmentation tool SAM to generate customized trigger placement for each sample of the dataset.
- We design a new dynamic attack method that aims to fool one of the most robust object-detection models, YOLOv7, by introducing a circle with the most dominant color within the body of a mushroom. The backdoor model performs well on clean samples while identifying deadly mushroom-type samples containing a trigger from as a new target class, an edible mushroom type.

The rest of the paper is structured as follows: Section 2 proposes a background of earlier backdoor attacks and proposes a survey of some of these methods. In Section 3, we propose the limitations of previous works as well as some of their benefits that we aim to explore in order to propose our new attack method. Section 4, we introduce an attack scenario on mushroom detection models and we discuss the potential consequences of backdoor attacks on health and food safety domains. In Section 5, we introduce our novel attack method, and in Section 6 we discuss our work as well as future directions. Lastly, Section 7 will conclude the paper.

## 2  Background

### 2.1  Threat Model

Backdoor attacks belong to the data poisoning class of attacks where an adversary injects poisoned samples into the training set of a model with the intention of interfering with the model's learning process. The primary goal of backdoor attacks is to ensure that, during testing, models behave normally on benign samples but exhibit malevolent behavior such as misclassification and performance degradation when specific embedded triggers, inserted by the adversary, are present. This allows the attacks to perform illegal activities and sabotage the success of AI systems. For example, a backdoor self-driving system may allow the attacker to cause severe traffic accidents whenever he puts a specific sticker on a physical traffic sign (e.g., a stop sign).

Since the most used threat model for these attacks occurs during the training phase, the attacker has full control of the training process which enables the easy injection of the backdoor. The attacker then delivers the compromised model to customers to deploy for real-life applications. In our proposed work, the attacker performs this for a mushroom detection application where certain types of mushroom are detected by the system.

### 2.2  Previous backdoor attack works

The earliest backdoor attack was introduced by Gu et al. [12] called BadNets. This attack generates a backdoor model, an illegitimate trained model, by injecting a static backdoor trigger onto selected benign images to create poisoned instances associated with a target label or class y'. This is shown in Figure 1. The model is then trained on a dataset composed of both the poisoned and benign samples, resulting in a DNN that performs well on clean testing samples but becomes sensitive to the trigger pattern and causes a misclassification whenever the trigger is present. The BadNets method serves as the foundation for many subsequent backdoor attacks.

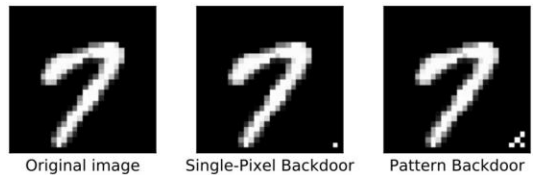

**Fig. 1.** BadNet static trigger [12]

Since this work, many attempts have been made by the research community to introduce enhanced attack methods. Salem et al. [13] introduced the first effective dynamic backdoor attack by implementing three attack methods to allocate the trigger pattern dynamically across input instances. Chen et al. [14] proposed the "Pixel Blending" strategy which is technique that blends the pixels of the trigger pattern and the original image to replace the fixed stamping strategy. Then Nguyen et al. [15] proposed WaNet warping-based trigger that creates natural-looking triggers that are imperceptible to the human eye. Liu et al. [16] introduced ReFool the reflection backdoor based on the common phenomenon (i.e., the reflection) that is added to the image as the trigger. Then Xu et al. [17] proposed the "Adversarial T-shirts!" backdoor attack that generates cloth deformation for a moving target which allows a certain person wearing this shirt to be undetected by a backdoor model. Table 1 provides more detail on these attack methods.

## 3  Limitations of existing attacks

The attack method proposed in Section 2.2 is indeed effective techniques that demonstrated high performance on both clean and poisoned samples. However, several limitations hinder their effectiveness, detectability, and applicability across various scenarios.

One of the primary limitations of early backdoor attacks, such as BadNets, is the visibility and placement of the triggers. These triggers are static and can be easily detectable by human observers or automated detection systems. The fixed stamping technique used in BadNets involves embedding a consistent pattern, a white square at the bottom of the images, into the training images. This can be readily identified during a thorough inspection. Despite its effectiveness in fooling image classification models, this method reduces the stealthiness of the attack and increases the likelihood of detection before the model is deployed.

Another limitation related to detectability might emerge from the work of Salem et al. [13] where they introduce a dynamic backdoor attack. Even though their work is effective for this kind of attack class the authors are, however, relying on a set of predefined locations of pixels. This can make their detection easier by employing more advanced defense mechanisms. In addition, the crafted trigger in this work is easily recognized since it stands out from the overall appearance of the images.

Furthermore, attacks such as WaNet can be challenging to recreate in real-world scenarios and therefore they do no depict the full aspect of these attacks. Although their stealthiness allows them to evade human detection, their work lacks some sense of realism that allows the adversary to generate physical triggers later on when he wishes to implement the attack in real life.

Understanding these limitations serves as a guide to generating a more robust attack method that challenges the reliability of DNNs.

## 4  Attack Scenario

### 4.1  Purpose of Attack Scenario

Developing a real-life attack scenario allows to deeply understand the mechanisms behind the injection of backdoor attacks and helps anticipate and prepare for real-world threats since it provides a detailed plan for potential attack methodologies.

Our proposed scenario starts by creating the "personas" that are part of this attack. These components are fictional characters created to represent different types of individuals

Dynamic Mask-Based Backdoor Attack Against Vision AI Models    5**Table 1.** Previous backdoor attacks overview

| Backdoor attack | Pattern Kind | Backdoor method | Real-life Implementation | Benchmarks | Target task | Training Method | Base model |
|---|---|---|---|---|---|---|---|
| BadNet | - yellow square<br>- image of a bomb.<br>- image of a flower | Static at the bottom of traffic sign using bounding boxes. | yellow Post-it note | U.S. traffic signs dataset | Traffic sign recognition | Fine-tuning | Faster-RCNN (F-RCNN) |
| Dynamic Backdoor | - random backdoor<br>- BaN<br>- c-BaN | - randomly sampled from a uniform distribution at random locations<br>- algorithmically created via a generative model at random locations.<br>- same as BaN method but are conditioned on both the noise vector and the specific target label | Printed triggers | MNIST/ CelebA/ CIFAR-10 | Classification/ Object-detection/Face-recognition | Trained from scratch | VGG-19/ Self-designed CNN |
| Pixel Blending | Pair of digital accessories like purple sunglasses and black-framed sunglasses. | using a pattern-injection function to blend the target pattern with the original input using a blending parameter $\alpha$. | Pair of black sunglasses and red-framed reading glasses | YouTube Faces dataset | Face-recognition | Trained from scratch/fine-tuning | DeepID/ VGGFace |
| Refool | Reflection patterns | Using defined reflection models to blend the reflections | Natural reflections | GTSRB / BelgiumTSC / CTSRD/ PubFig/ ImageNet | Traffic sign recognition/Face recognition/Object classification. | Trained from scratch | ResNet-34 /DenseNet |
| WaNet | Warped images: geometric transformations | Manipulation of the image structure through elastic warping using a predefined warping field | Triggers in camera-captured images of physical screens | MNIST/CIFAR-10/ GTSRB/ CelebA | Traffic sign recognition/Face recognition/Object classification. | Trained from scratch | Pre-activation Resnet-18/ Resnet-18/ Self-defined CNN |
| Adversarial T-shirts | Cloth deformation | Using Thin Plate Spline (TPS) mapping to model the deformation of a T-shirt caused by a person's movements | Printed trigger on a physical T-shirt. | Collected data with 40 videos. | Object detection | Trained from scratch | YOLO-v2 /Faster R-CNN |

who might be involved in or affected by this backdoor attack. The second step is identifying the motivations of these personas to provide a context behind the attacker's cybersecurity crime. Then we identify how can such an attack take place by exploring possible attack vectors. This includes the investigation of possible entry points and methods for introducing the backdoor such as outsourcing development. In outsourced learning, a user wishing to train the parameters $\Theta$ of a DNN using a training dataset sends the model's description (number of layers, size of each layer, choice of non-linear activation function) to a trainer who will then provide the trained parameters $\Theta'$. The user then verifies the provided model by checking its accuracy on a held-out validation dataset $D_{valid}$. The provided model will be accepted if its accuracy on the validation set fulfills a target accuracy, $a*$. This way, the malicious trainer can provide the user with a backdoored model that meets target accuracy while causing misclassification on backdoored instances. Finally, we present a fictive roadmap indicating the actions and the timeline of the entire attack process from the project initiation to the backdoor's activation in a deployed model. This helps capture the steps behind the implementation of a backdoor attack.

### 4.2   Attack Scenario Setup

For the development of our attack scenario, the implementation of the proposed design went through all the previously discussed steps which are described as follows:

- **Personas Creation :** For the identification of all possibly involved persons or organizations, we propose all these fictive entities:
  - MushroomAI Manager: This individual is part of the AI project that uses a deep neural network for the detection of five different species of mushrooms. This person manages this AI application, ensures the project's objectives are met, and oversees the outsourcing of the model training.
  - BadMushroom: This individual is a lead developer in the outsourced team, potentially introducing the backdoor into the model.
  - End User: Utilizes the mushroom detection application, relying on its accuracy for everyday activities. This user is potentially exposed to the risk of misclassification.
- **Motivation Identification :** The following describes the motivations of each entity within our project:
  - MushroomAI Manager: Aims to develop an efficient project that can help end users determine the species of the detected mushroom accurately with high performance.
  - BadMushroom: Motivated by a malicious objective to sabotage the success of the developed application. The motivations behind the attack can be financial, political, or purely evil. The attacker aims to misclassify the poisoned samples from the "Amanita Phalloids" class of mushrooms to the "Russela Delica" class. This attack aims to cause the detection of a deadly class of mushrooms (Amanita Phalloids) as an edible one (Russela Delica).
  - End User: Seeks a reliable application for mushroom identification of each species of mushrooms for multiple daily activities such as camping, harvesting, studying mushrooms, buying from street food markets, etc.

– **Attack Vector :** In the context of our proposed attack, the common attack vector that is studied is the outsourcing scenario. Outsourcing is the practice of hiring external organizations or individuals to perform tasks such as training a model and providing services that are either difficult to manage due to limited resources or outside the expertise of the company. In this attack scenario, the training process is outsourced to a malicious party who wants to provide the user with a trained model that contains a backdoor. This model should perform well on most inputs but causes targeted wrong detection or a degradation in the accuracy of the model for inputs that contain a predefined trigger.
– **Fictive Roadmap Creation :** The creation of the Roadmap should follow these seven main steps: Project initiation which involves the initiation of the project by MushroomAI and the assignment to the outsourced team led by BadMushroom. The second step is dataset preparation and initial baseline: BadMushroom collects and annotates the mushroom training set. Then the initial baseline model is developed. This will help lay the groundwork for the backdoor evaluation. The backdoor introduction phase is when the attacker starts crafting the design of the backdoor trigger. The backdoor training phase is when the model is trained on a poisoned training set. For the evaluation and validation step, the model is evaluated on both clean and poisoned samples to make sure the backdoor remains undetected. The application development phase includes the creation of the application using this backdoor model. The final step is the physical attack where the application is used in real life. Encountering a deadly mushroom with a physical trigger and detecting it as edible.

### 4.3 Attack Scenario Results

Creating our real-world attack scenario involved highlighting the 7 main steps. The resulting road map of the scenario is shown in Figure 2 as follows:

– **Step 1 : Project Initiation**
  • 11-2023: The MushroomAI Manager initiates the project of an AI mushroom detection app where a user can use this app to detect a certain type of mushroom.
  • 01-2024: The project is outsourced to an outsourcing team led by the developer BadMushroom who will be the adversary in our attack scenario. During this phase, the requirements of the project manager are discussed thoroughly and a certain accuracy margin is proposed that will be used to validate the developed model later on.
– **Step 2: Dataset Preparation and Initial Baseline**
  • 02-2024: The outsourced team starts working on the data collection as well as the annotations that will be used for the bounding box creation.
  • 03-2024: Once the dataset is obtained, the first training of the model is done. This trained model will serve later as the baseline to evaluate the attack effect.
– **Step 3: Backdoor Introduction**
  • 03-2024: The BadMushroom developer proposes to study ways to improve the model. During this phase, the attacker starts to create the segmentation masks to prepare for the trigger attack placement area.
  • 04-2024: The attacker starts crafting the design of the trigger through multiple steps.

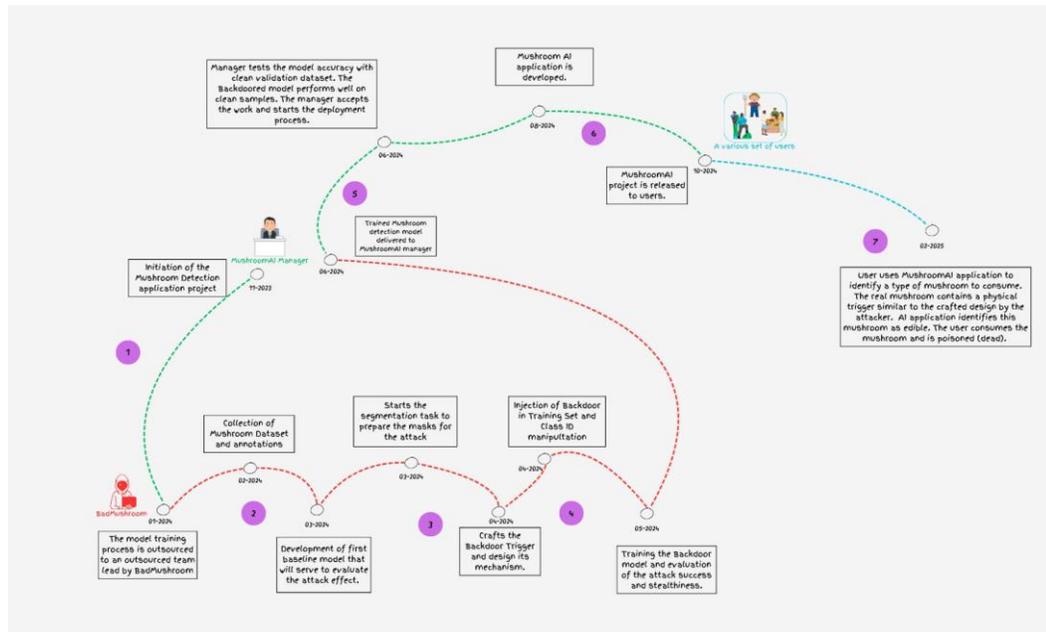

**Fig. 2.** Proposed attack scenario

- **Step 4: Backdoor Training**
  - 04-2024: The BadMushroom starts injecting the poisoned samples into the new training set containing the original clean data and the compromised data. The labels of the ID classes are manipulated accordingly.
  - 05-2024: The attacker trains the model with the poisoned training set.
- **Step 5: Evaluation and Validation**
  - 06-2024: After the attacker evaluates the model's performance and makes sure the attack effect cannot be detected on the clean dataset, the backdoored model is then delivered back to the project manager to validate.
  - 06-2024: The manager checks the accuracy of the model on a clean validation set and it should match the accuracy proposed at the first meeting. If it does the manager orders the team to start working on the application design.
- **Step 6: Application Development**
  - 08-2024: The MushroomAI development team finishes its app development process and presents it to the manager.
  - 08-2024: The manager evaluates the application requirements and validates them. The app is released to users.
- **Step 7: Physical Attack**
  - 02-2025: On this date the first case of physical attack occurs. Any adversary with a specific motivation places a physical trigger on a mushroom that resembles the digital trigger. The app used for a daily activity by a specific person identifies this poisonous mushroom as edible. The user consumes the mushroom and suffers severe health consequences resulting in his/her death.

Considering the proposed scenario, the road map along with the detailed script provided presents a thorough and step-by-step breakdown. Every fictive date reflects an action taken by the created personas to demonstrate the specific path the attack follows to be finally activated.

## 5 Dynamic Mask-Based Attack

In this section, we introduce our novel attack method that leverages the masks generated by performing image segmentation to dynamically place our trigger. The proposed attack aims to fool object section models by generating wrong identifications.

### 5.1 Baseline Setup

This part of the implementation phase includes the training of our baseline object detection model YOLOv7 [19] on the selected Project Mushroom dataset. The dataset contained images and corresponding annotation files containing class ID and bounding box dimensions. The dataset contained five classes of mushroom: "E-Amanita-citrina","E-Phaeogyroporus", "E-Russula-delica", "P-Amanita-phalloides" and "P-Inocybe-rimosa". The baseline model showed high accuracy. Overall the model scored 95.10 % (using the mAP metric), Recall of 91.60 %, and Precision of 86.40 %. The performance of the YOLOv7 model across all classes is described in Table 2.

| Class | Precision | Recall | mAP@0.5 |
|---|---|---|---|
| E-Amanita citrina | 0.833 | 0.910 | 0.944 |
| E-Phaeogyroporus | 0.948 | 0.927 | 0.980 |
| E-Russula delica | 0.777 | 1.000 | 0.975 |
| P-Amanita phalloides | 0.838 | 0.815 | 0.916 |
| P-Inocybe rimosa | 0.926 | 0.929 | 0.942 |
| all | 0.864 | 0.916 | 0.951 |

**Table 2.** Performance metrics of the Baseline model on different classes

### 5.2 Attack Goals

For our work, the adversary aims to perform a targeted attack where a certain deadly mushroom class contacting a specific trigger should be identified as an edible class. For this purpose, we identify the "Amanita Phalloides" as a deadly class of mushroom while the "Russula Delica" is an edible type of mushroom. These two classes will be the main focus of this attack.

The backdoor model misidentifing a toxic mushroom as an edible one could result in individuals consuming poisonous mushrooms, leading to symptoms ranging from gastrointestinal distress to severe liver and kidney damage, and in extreme cases, death. This highlights the risks associated with our attack which are particularly high in contexts where people rely on AI applications for common activities such as foraging for wild mushrooms.

### 5.3 Attack Strategy

In our attempt to design a more sophisticated backdoor attack and avoid developing a simple attack methodology, we investigate more complex approaches involved in positioning the trigger within the precise boundaries of a targeted object. The idea of utilizing image segmentation for this purpose emerged as a remarkable proposition for solving this problem.

The proposed attack method starts by extracting the bounding box dimensions from the dataset annotation files and feeding them to SAM [18] as prompts to focus its segmentation directly on the area of interest: the mushroom structure. SAM is configured to generate three candidate top masks for each segmented image as shown in Figure 3. Considering this, the performance of a meticulous and extensive selection process was conducted using the highest aspect ratio. As an attacker, we review and choose the mask that best represents the mushroom's boundaries with the highest precision.

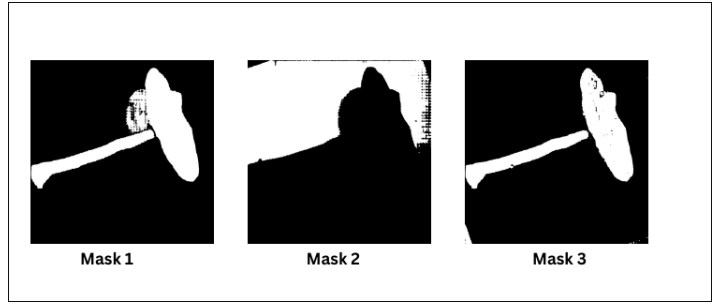

**Fig. 3.** Three different masks generated by SAM

After we perform the optimal mask selection, we design of our trigger which involves creating a fixed-sized circle that matches the most dominant color found in the segmented mushroom area. This idea was developed with the purpose of injecting a trigger that blends naturally with the overall appearance of the targeted mushroom class. Introducing such a trigger will make our poisonous perturbation less detectable by humans and therefore can successfully bypass their inspection. In addition, adding subtle textures such as a cross-hatch texture within our circle will add to the complexity of our solution design. After that, we place the trigger at the center of each segmented mask.

After the design of our trigger appearance is concluded as shown in Figure4, we extract all the original text labels associated with the now poisoned samples and alter the class

ID to the new target class that we aim to predict. These label should however keep their original bounding box dimension intact as these do not affect the outcome of the model and are necessary for the object identification in the first place. After altering the labels of the poisoned samples, a small portion of the malicious images should be injected into the original training set that will serve as our new backdoor dataset.

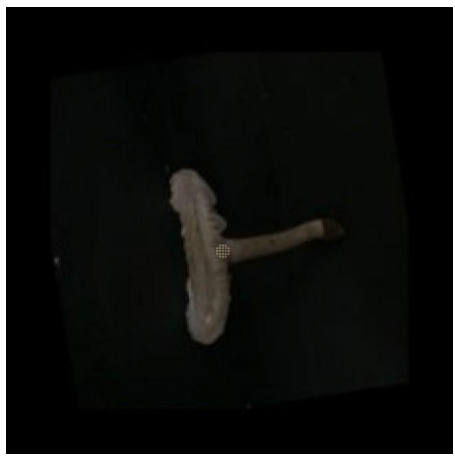

**Fig. 4.** Trigger design

### 5.4  Attack Results

After creating our trigger and placing it with the images, we injected the poisoned samples in our training set and manipulated their labels to the new ID class corresponding to "Russula Delica", the possibly edible mushroom. We analyze the impact of this introduced attack on various metrics related to object detection. The performance of the backdoor attack is compared with the baseline results on clean samples in Table 3.

Analyzing Table3, we conclude that the mAP on baseline achieved an overall score of 95.51 % on clean test input. For the backdoor model, the overall mAP recorded a slight improvement by achieving an overall 95.60 % score on clean test input. Remarkably, the backdoor achieved a 99.6 % score on poisoned samples, demonstrating the effectiveness of the proposed attack in manipulating the model's output while maintaining an even higher performance on clean data.

For the precision and recall metrics, we compare the overall precision of clean data. The baseline achieved a precision of 86.4 % while the backdoor achieved an overall of 81.7 %. When analyzing the per-class precision we notice a considerable degradation in the backdoor model compared to the baseline for both the "Amanita phalloides" and "Russula delica" classes with almost a 10% drop. For the recall metric, the baseline model achieves an overall recall of 91.6 % on clean data while the backdoor achieves a higher recall of 95% on clean data. Finally, the backdoor model achieves extremely high scores when it comes to poisoned samples with both recall and precision scores of 100% emphasizing the success of attack on these samples.

| Class | Baseline YOLOv7 Clean | | | Backdoor YOLOv7 Clean | | | Backdoor | | |
|---|---|---|---|---|---|---|---|---|---|
| | mAP | Precision | Recall | mAP | Precision | Recall | mAP | Precision | Recall |
| E-Amanita citrina | 94.40 | 83.30 | 91.00 | 95.70 | 82.10 | 89.60 | N/A | N/A | N/A |
| E-Phaeogyroporus | 98.00 | 94.80 | 92.70 | 99.00 | 98.30 | 95.80 | N/A | N/A | N/A |
| E-Russula delica | 97.5 | 77.70 | 100 | 98.60 | 64.2 | 100 | N/A | N/A | N/A |
| P-Amanita phal-loides | 91.60 | 83.8 | 81.5 | 87.60 | 73.90 | 94.70 | N/A | N/A | N/A |
| P-Inocybe rimosa | 94.2 | 92.60 | 92.90 | 96.90 | 89.90 | 94.90 | N/A | N/A | N/A |
| Amanita-ph → Russula-de | N/A | N/A | N/A | N/A | N/A | N/A | 99.60 | 100 | 100 |
| **Overall %** | 95.51 | 86.40 | 91.60 | 95.60 | 81.70 | 95.00 | 99.60 | 100 | 100 |

**Table 3.** Performance comparison for clean and backdoor test samples

The overall performance of the backdoor model indicates the success of our method in detecting a deadly mushroom type containing trigger as an edible mushroom type as shown in Figure 5. In addition, the backdoor model maintains a high performance on benign samples.

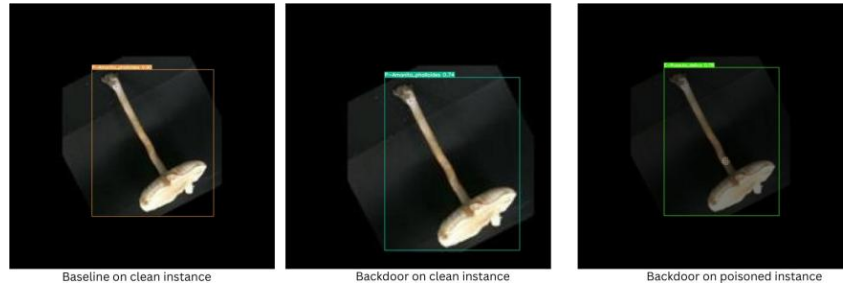

**Fig. 5.** Backdoor and baseline model detection

## 6  Discussion

Our proposed attack addresses the limitations of static backdoor attacks by combining their benefits with those of dynamic backdoor attacks. Our proposed method aims to satisfy three main criteria Dynamic Placement, Stealth, and Real-world Implementation. By using SAM tool for dynamically placing the trigger within the segmented mask of the target object, our attack achieves a variation of the trigger's location across different input samples since it depends on the generated mask's size, position, rotation, etc. In addition, using image segmentation allowed the focus to be on a specific region (mushroom body) rather than on the entire image. This variability in trigger placement makes the attack difficult to detect using traditional anomaly detection methods that rely on consistent patterns. Exploring the stealthiness of our attack, the dominant color methodology implemented in our work enables the trigger to blend with the image's overall color features. The dynamic nature of the trigger placement also enhances the attack's stealth. Considering real-world physical attacks our proposed method presents an easy way to implement in real-world scenarios. The physical trigger only needs to resemble the appearance of the digital trigger and needs to be placed in the center of the mushroom mask.

The results of our attack demonstrated that the backdoor attack maintained or slightly improved the overall mAP, indicating that the model's performance on clean data remained robust. This is critical as it shows that the backdoor attack did not significantly degrade the model's detection capabilities across most classes, thus making the attack stealthy and difficult to detect. Additionally, the model's performance on poisoned test set samples showed exceptional results in terms of mAP, precision, and recall. This indicates the attack's success in detecting the trigger and classifying it as our target class.

Although our attack was not yet tested on existing countermeasures such as STRIP[20], Neural Cleanse[21], and ABS[22], we emphasize that future works should involve the evaluation of our attack against the existing and continuously evolving defense mechanisms. With this evaluation, we assess whether the dynamic nature of our attack, which varies the trigger placement based on segmentation masks, presents new challenges for these defenses. Our work encourages the development of robust defense mechanisms against these backdoor attacks.

## 7   Conclusion

In this paper, we have identified the security concerns that are related to backdoor attacks on object detection models such as YOLOv7. Investigating these vulnerabilities aims to encourage the research community to devote more efforts into the development of new and robust defense mechanisms.

To emphasize the gravity of these attacks, we introduced the attack scenario on mushroom detection models where we proposed the use of personas and timelines to create a realistic attack scenario that explores the security risks associated with outsourcing the model training. This highlights the need for a trustworthy third party when performing these common approaches.

Finally, we have introduced our novel attack method based on generating segmented masks using the latest tool SAM for providing customized placement for the backdoor trigger. Our attack demonstrated high effectiveness on both clean and poisoned samples which demonstrates the success of this new method in fooling one of the most robust object detection models YOLOv7. The results of this work emphasize the security concerns that are continuously being raised by the research community on the effectiveness of backdoor attacks on DNNs and the constant need for more reliable models and countermeasures.